%% file: sample-sigconf-authordraft.tex
\documentclass[sigconf]{acmart}
\usepackage{algorithm}
\usepackage{algpseudocode}
\usepackage{colortbl}
\usepackage{multirow}
\usepackage{subfig}
\input{settings}
\AtBeginDocument{%
  }

\setcopyright{cc}
\setcctype{by} 
\copyrightyear{2026}
\acmYear{2026}
\acmDOI{10.1145/3767308.3836460}
\acmConference[MM '26]{Proceedings of the 34th ACM International Conference on Multimedia}{November 10--14, 2026}{Rio de Janeiro, Brazil}
\acmBooktitle{Proceedings of the 34th ACM International Conference on Multimedia (MM '26), November 10--14, 2026, Rio de Janeiro, Brazil}
\acmISBN{979-8-4007-2213-4/2026/11}

\begin{document}

\title{SHARP: Spectrum-aware Highly-dynamic Adaptation for Resolution Promotion in Remote Sensing Synthesis}

\author{Bingxuan Zhao}
\email{bxuanzhao202@gmail.com}
\orcid{0009-0002-3045-6175}
\affiliation{%
  \institution{Northwestern Polytechnical University}
  \city{Xi'an}
  \country{China}
}

\author{Qing Zhou}
\email{chautsing@gmail.com}
\orcid{0009-0004-3300-0805}
\affiliation{%
  \institution{Northwestern Polytechnical University}
  \city{Xi'an}
  \country{China}
}

\author{Chuang Yang}
\email{omtcyang@gmail.com}
\orcid{0000-0002-4086-8387}
\affiliation{%
  \institution{The Hong Kong Polytechnic University}
  \city{Hong Kong}
  \country{China}
}

\author{Junyu Gao}
\email{gjy3035@gmail.com}
\orcid{0000-0001-6000-8168}
\affiliation{%
  \institution{Northwestern Polytechnical University}
  \city{Xi'an}
  \country{China}
}

\author{Qi Wang}
\authornote{Corresponding author.}
\email{crabwq@gmail.com}
\orcid{0000-0002-7028-4956}
\affiliation{%
  \institution{Northwestern Polytechnical University}
  \city{Xi'an}
  \country{China}
}

\renewcommand{\shortauthors}{Zhao et al.}

\begin{abstract}
Text-to-image synthesis for remote sensing (RS) lacks an accessible, high-performance generative foundation, as directly training diffusion models at large, high resolutions is computationally prohibitive. Training-free resolution promotion via Rotary Position Embedding (RoPE) extrapolation offers an efficient alternative, but existing algorithms apply static scaling rules tailored to natural scenes, whereas RS imagery is dominated by dense, minute instances that hinge on high-frequency structural integrity. We present a comprehensive framework for large-scale RS synthesis. First, we curate over 100{,}000 image-text pairs to train RS-FLUX, a domain-specialized generative prior. Second, we propose \textbf{SHARP}, a training-free extrapolation algorithm that introduces a Rational Decay Scheduler to modulate RoPE frequencies throughout denoising: strong positional extrapolation early on enforces coherent global layouts, and progressive relaxation later recovers dense high-frequency details. Extensive experiments show that SHARP consistently achieves state-of-the-art performance across multiple promoted resolutions with negligible ($<$4\%) overhead, offering an efficient and structurally faithful solution for large-scale RS generation. Code is available at \url{https://github.com/bxuanz/SHARP}.
\end{abstract}





\maketitle

\begin{figure}[t]
  \centering
  \includegraphics[width=0.95\linewidth]{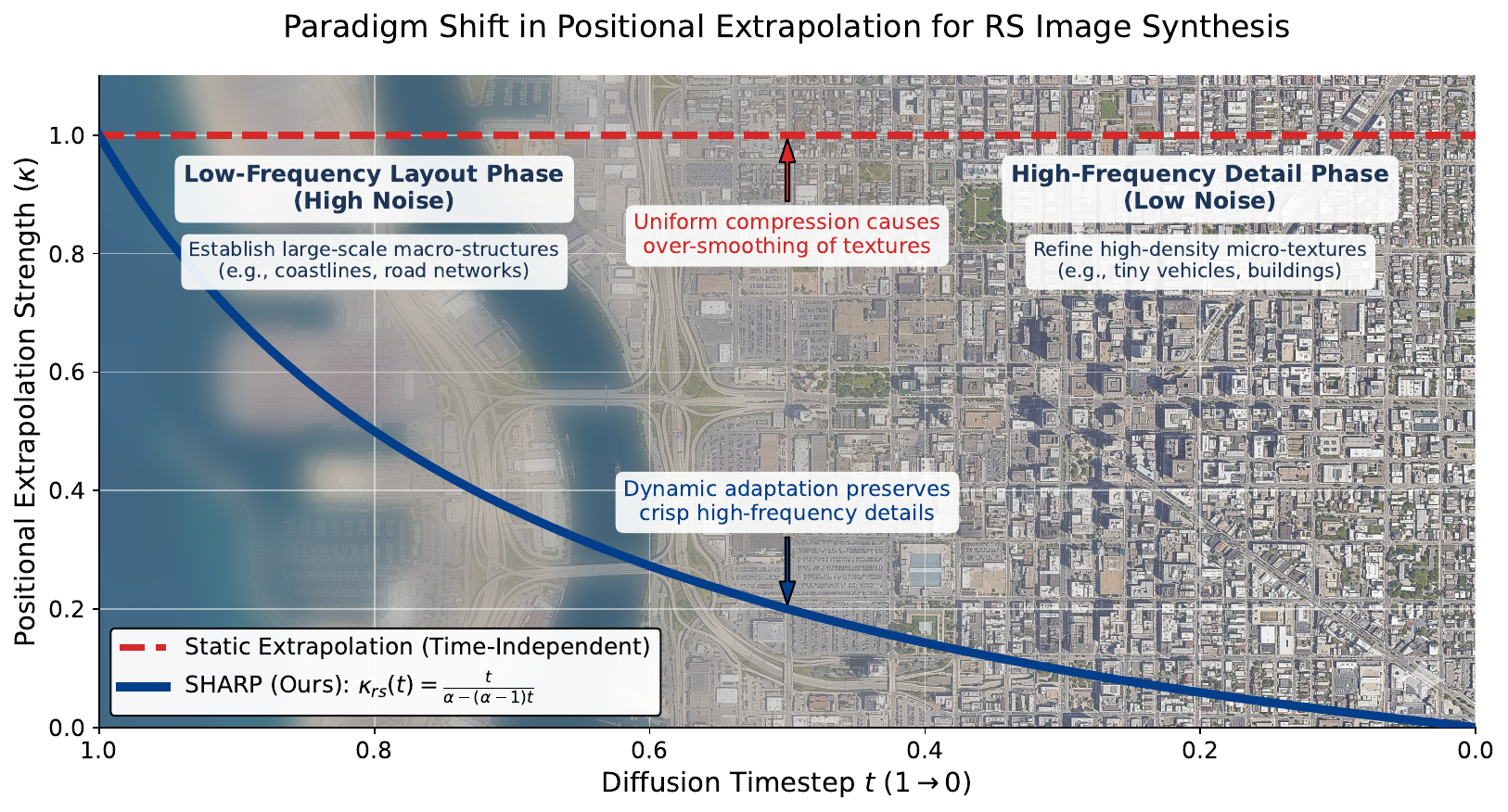}
  \caption{\textbf{Static vs.\ dynamic positional extrapolation for resolution promotion (conceptual illustration).} Diffusion-based generation follows a frequency-progressive paradigm: global layouts emerge at high noise levels while dense high-frequency details crystallize at low noise levels. Static methods (red dashed) apply uniform extrapolation strength across all steps, uniformly compressing the spatial spectrum and causing catastrophic over-smoothing of fine remote sensing structures. \textbf{SHARP} (blue solid) achieves highly-dynamic adaptation by coupling the extrapolation strength with the denoising progression via a Rational Decay Scheduler $\kappa_{rs}(t)$, continuously preserving crisp high-frequency details without sacrificing structural coherence.}
  \Description{A line graph comparing static and dynamic positional extrapolation, showing SHARP preserving high-frequency details across denoising steps, alongside visual examples of remote sensing images.}
  \label{fig:teaser}

\end{figure}

\section{Introduction}
\label{sec:intro}

Remote sensing (RS) imagery \cite{amin2009special,goktepe2025ecomapper, soille2002advances, wu2021convolutional} is central to Earth observation tasks such as scene classification \cite{jia2025few}, semantic segmentation \cite{jia2025entity,10847788}, and multi-view reconstruction \cite{10720843,11112633,11595022}, yet the community lacks an accessible, high-performance generative foundation for text-to-image synthesis \cite{tang2024crsdiffcontrollableremotesensing, yu2024metaearthgenerativefoundationmodel, tang2025aerogenenhancingremotesensing}. A critical bottleneck in advancing this field is the inherent demand for large-scale, high-resolution imagery. A single generated RS image must span a vast spatial extent while preserving a massive amount of dense, scattered, and tiny local structures (e.g., vehicles, complex road networks) \cite{li2020object,ou2025geopixmultimodallargelanguage}. Directly training foundational Diffusion Transformers (DiTs) \cite{peebles2023scalable} at such extreme resolutions incurs prohibitive computational costs due to the quadratic scaling of self-attention. Therefore, achieving training-free \emph{resolution promotion} at inference time---generating at sizes vastly beyond the training resolution (e.g., $1024{\times}1024 \!\to\! 2048{\times}2048$)---is highly desirable.

Current training-free resolution promotion approaches predominantly rely on \emph{positional extrapolation}, specifically by rescaling Rotary Position Embedding (RoPE) \cite{su2024roformer} frequencies (e.g., PI \cite{chen2023extending}, NTK-aware \cite{bloc97ntk2023}, YaRN \cite{peng2024yarn}). However, these algorithms were designed for Large Language Models (LLMs) or natural images, which typically feature single salient subjects and smooth textures. In stark contrast, RS imagery is overwhelmingly dominated by high-frequency structural details. Directly migrating these methods to RS generation exposes a severe domain mismatch.

Crucially, existing training-free algorithms rely on a \emph{static} extrapolation rule that remains fixed throughout denoising (red dashed line in Fig.~\ref{fig:teaser}). This rigid assumption conflicts with the coarse-to-fine, frequency-progressive nature of diffusion models \cite{ho2020denoising,song2020score, zhang2023adding, podell2024sdxl, rombach2022high, du2024demofusion}: the early high-noise phase converges on the low-frequency \emph{global layout} (e.g., broad coastlines, primary road networks), while the late low-noise phase crystallizes high-frequency \emph{texture details} (e.g., dense vehicles, minute building edges). Although strong compression is necessary early on to establish an expanded layout, it becomes a severe spectral bottleneck during detail refinement---continuously compressing spatial frequencies precisely when delicate micro-textures are trying to emerge. For RS imagery, this irreparably suppresses high-frequency components, causing catastrophic over-smoothing that washes out the structural integrity of fine instances.

To overcome this fundamental domain mismatch and computational bottleneck, we present a comprehensive framework for large-scale RS text-to-image synthesis. First, we establish a robust domain-specialized prior (\textbf{RS-FLUX}) by fine-tuning FLUX \cite{flux2024} on a curated dataset of over 100{,}000 high-quality RS images. Second, to accomplish high-fidelity resolution promotion, we propose \textbf{SHARP} (\textbf{S}pectrum-aware \textbf{H}ighly-dynamic \textbf{A}daptation for \textbf{R}esolution \textbf{P}romotion). 

Unlike static baselines, SHARP achieves \emph{highly-dynamic adaptation} by introducing a Rational Decay Scheduler $\kappa_{rs}(t)$ (blue solid line in Fig.~\ref{fig:teaser}) that continuously modulates RoPE frequencies step-by-step. Our core insight is to explicitly synchronize the positional extrapolation strength with the natural spectral evolution of the diffusion process. During the early global-layout generation phase, SHARP applies strong positional extrapolation to guarantee macro-layout coherence across the expanded spatial canvas. As the generation transitions into the texture-refinement phase, SHARP smoothly and continuously decays the extrapolation strength. By deliberately reducing the positional compression when precise textures are forming, SHARP liberates the model to render crisp micro-structures without spatial distortion. This dynamic decoupling seamlessly expands the image footprint while faithfully preserving intricate RS details.

In summary, our main contributions are:
\begin{itemize}
    \item \textbf{Domain-Specialized Generative Prior:} We construct a 102{,}952-pair RS text-to-image dataset and fine-tune FLUX to build \textbf{RS-FLUX}, providing a powerful foundational DiT for the RS community.
    \item \textbf{Identifying the Static Bottleneck:} We identify and theoretically analyze the fatal conflict between existing \emph{static} positional extrapolation and the frequency-progressive nature of diffusion, revealing the root cause of over-smoothing in high-frequency RS generation.
    \item \textbf{Dynamic Extrapolation Algorithm (SHARP):} We propose \textbf{SHARP}, a novel training-free algorithm that dynamically couples RoPE frequencies with the denoising spectrum. Extensive experiments demonstrate that SHARP effectively eliminates over-smoothing, achieving state-of-the-art performance across multiple promoted resolutions with negligible overhead.
\end{itemize}

\section{Related Work}

\subsection{Remote Sensing Image Generation}
Generative modeling for RS imagery has evolved significantly, transitioning from GAN-based methods to highly expressive diffusion models \cite{ho2020denoising, he2024scalecrafter}. While early RS generative research demonstrated the value of diffusion priors, they predominantly targeted specific restoration or conditional tasks (e.g., super-resolution \cite{saharia2021imagesuperresolutioniterativerefinement}, cloud removal \cite{sui2024diffusionenhancementcloudremoval}, SAR-to-RGB translation \cite{11178091}, and object-centric generation \cite{10922144}). For open-ended text-to-image synthesis, most existing RS adaptations continue to rely on U-Net backbones \cite{sebaq2024rsdiff}, which are inherently biased toward patch-level synthesis rather than holistic, complex scene generation. The recent success of Diffusion Transformers (DiTs) \cite{peebles2023scalable, flux2024} in producing photorealistic natural images by scaling self-attention motivates our development of \textbf{RS-FLUX}, a DiT-based generative foundation model tailored for the dense and complex structural statistics of RS imagery.

\subsection{Positional Extrapolation in Diffusion Transformers}
\label{sec:rope_extrapolation}
Training foundational DiTs at extreme resolutions is computationally prohibitive. Consequently, training-free resolution promotion---generating images at scales vastly exceeding the training context window---has become essential \cite{zhaosigma}. In modern DiTs, Rotary Position Embedding (RoPE) \cite{su2024roformer} is the standard spatial encoding mechanism. For a token at position $m$, RoPE encodes positional information by rotating the hidden representations using a set of fixed frequencies:
\begin{equation}
\theta_i = b^{-2i/d}
\end{equation}
where $d$ is the feature dimension, $i \in [0, d/2)$ is the channel index, and $b$ is the frequency base (typically 10000). The corresponding rotation angle applied to the feature space is $m\theta_i$. 

When promoting resolution at inference time, the positional index $m$ exceeds the context window seen during training, leading to severe out-of-distribution (OOD) degradation. Representative training-free methods mitigate this by rescaling the position or frequencies. Position Interpolation (PI) \cite{chen2023extending} linearly compresses positions to $m/s$ (where $s$ is the resolution promotion factor), effectively compressing all frequencies uniformly. NTK-aware scaling \cite{bloc97ntk2023} modifies the base wavelength $b$ to preserve high-frequency information, while YaRN \cite{peng2024yarn} applies a more sophisticated, frequency-selective ramp. While computationally lightweight, these extrapolation techniques were originally designed for Large Language Models (LLMs) or natural images, overlooking the unique structural demands of the RS domain.

\subsection{The Bottleneck of Static Extrapolation}
A fundamental limitation shared by PI, NTK-aware scaling, and YaRN is their \emph{static} strategy: the same frequency modification is applied to $\theta_i$ at every denoising timestep $t$, ignoring the frequency-progressive generative physics of diffusion \cite{ho2020denoising,song2020score}. For high-frequency-dominated RS imagery, compressing the positional spectrum $m\theta_i$ in the late denoising stages suppresses exactly the components required for resolving fine instances, inevitably yielding over-smoothed outputs (formally analyzed in Sec.~\ref{sec:spectral_formulation}). SHARP breaks this static assumption with a dynamically modulated promotion schedule that respects the instantaneous spectral state of the diffusion process.

\section{Methodology}

\begin{figure*}[t]
    \centering
    \includegraphics[width=0.98\linewidth]{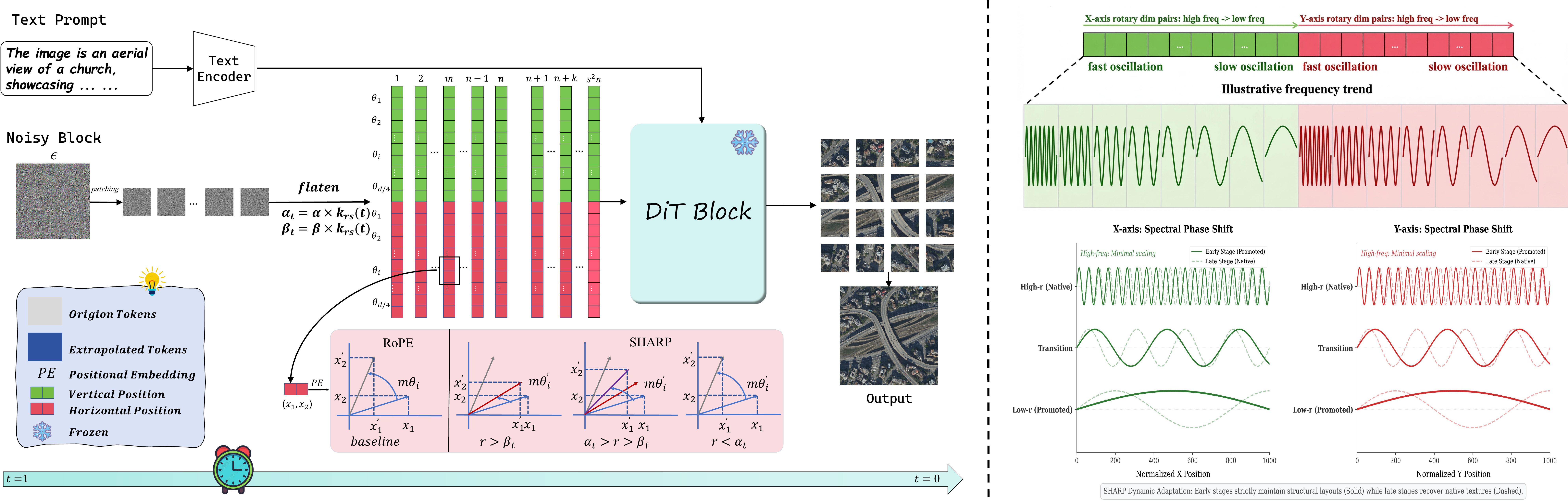}
    \caption{\textbf{SHARP Overview.} SHARP performs dynamic resolution promotion in RoPE through the Rational Decay Scheduler $\kappa_{rs}(t)$ and the ramp function $\gamma(r,t)$. Strong promotion is preserved in the early layout stage and progressively reduced in the late detail-recovery stage, avoiding the over-smoothing caused by static extrapolation.}
    \label{fig:sharp_structure}
    \Description{A flowchart illustrating the SHARP dynamic resolution promotion mechanism, featuring the Rational Decay Scheduler and the Dynamic Ramp Function modifying the RoPE frequency table.}
\end{figure*}

This section presents our two-component framework. Section~\ref{sec:rsflux} describes RS-FLUX, the domain-specialized DiT prior obtained by fine-tuning FLUX \cite{flux2024} on curated RS data. Section~\ref{sec:spectral_formulation} establishes the spectral motivation that directly informs the design of SHARP (Sec.~\ref{sec:sharp}), our training-free dynamic resolution promotion strategy. The overall pipeline is illustrated in Fig.~\ref{fig:sharp_structure}.

\subsection{Domain-Specialized Prior: RS-FLUX}
\label{sec:rsflux}

Open-source DiT checkpoints \cite{peebles2023scalable} are optimized for natural images and fail to capture the semantics and spectral statistics of aerial scenes. RS imagery contains dense man-made structures, small objects, and repetitive textures under-represented in natural-image priors \cite{deng2009imagenet}. Parameter-efficient approaches such as LoRA \cite{hu2022lora} constrain updates to a low-rank subspace that may be insufficient for bridging the large distribution gap between natural and RS imagery. We therefore perform full fine-tuning of FLUX \cite{flux2024} on over 100{,}000 high-quality RS images, updating all transformer parameters so that attention heads, feed-forward layers, and the final projection jointly adapt to the spectral and geometric statistics unique to overhead imagery. The resulting \textbf{RS-FLUX} serves as the base generator in all experiments. Because SHARP operates exclusively at inference time by modifying the RoPE frequency table, it requires no additional high-resolution data or retraining---the two components are fully decoupled.

\begin{figure}[t]
    \centering
    \includegraphics[width=0.95\linewidth]{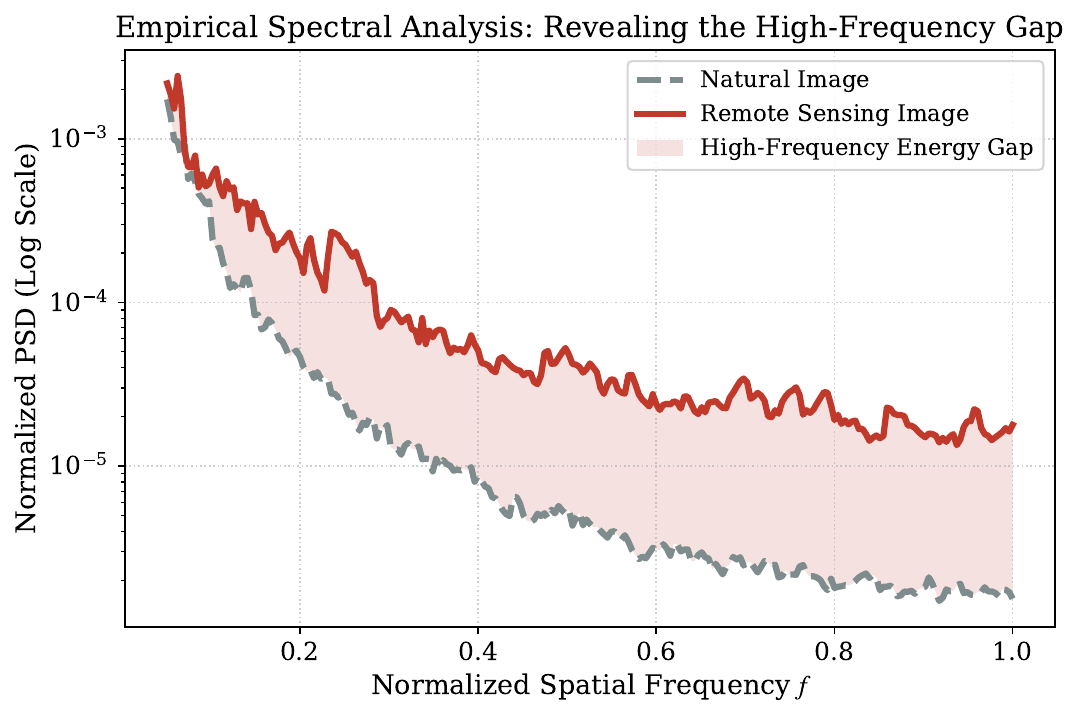}
    \caption{\textbf{Empirical Spatial Spectrum Analysis.} Average normalized radial power spectral density (PSD) over 100 ImageNet \cite{deng2009imagenet} and 100 DIOR \cite{li2020object} images. Each image is converted to grayscale, resized to $512\times512$, windowed with a 2-D Hanning function, transformed via centered FFT, radially averaged, and normalized by its non-DC low-frequency baseline. RS imagery exhibits systematically stronger medium- and high-frequency energy.}
    \label{fig:empirical_psd}
    \Description{A line graph comparing the radial power spectral density of ImageNet and DIOR datasets, showing remote sensing images have systematically stronger medium and high-frequency energy.}
\end{figure}

\begin{figure}[t]
    \centering
    \includegraphics[width=0.98\linewidth]{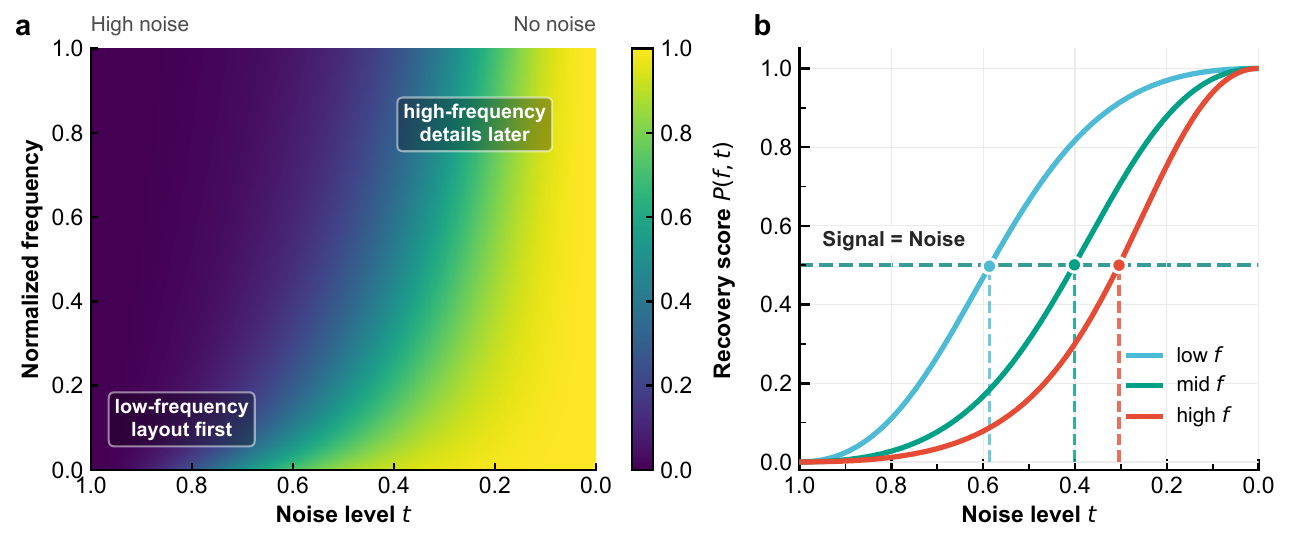}
    \caption{\textbf{Frequency-Progressive Denoising.} Left: heatmap of the bounded recovery score $P(f,t)=\rho(f,t)/(1+\rho(f,t))$ over normalized frequency and reverse denoising progress. Lower-frequency bands enter the signal-dominant regime earlier. Right: representative trajectories for low-, medium-, and high-frequency components, whose threshold crossings occur progressively later, motivating strong early promotion and weaker late extrapolation.}
    \Description{A heatmap displaying the bounded recovery score over normalized frequency and reverse denoising progress, and a line graph showing representative trajectories for different frequency components.}
    \label{fig:spectral_dynamics}
\end{figure}

\subsection{Spectral Motivation and Frequency-Progressive Denoising}
\label{sec:spectral_formulation}

To motivate the dynamic promotion strategy, we first quantify the spectral gap between natural and RS images. Following the protocol in Fig.~\ref{fig:empirical_psd}, we compute the average normalized radial PSD over 100 ImageNet \cite{deng2009imagenet} and 100 DIOR \cite{li2020object} images. After low-frequency normalization, RS imagery retains systematically stronger medium- and high-frequency energy (Fig.~\ref{fig:empirical_psd}). Uniform spectral compression is therefore more damaging for RS than for natural-image generation.

This empirical analysis establishes \emph{what} differs between the two domains; we next analyze \emph{when} different frequency bands become recoverable. Following the flow-matching formulation \cite{lipman2023flowmatchinggenerativemodeling} used by FLUX \cite{flux2024}, the noisy latent at normalized time $t \in [0,1]$ is
\begin{equation}
    x_t = (1-t)x_0 + t\epsilon,
    \label{eq:mixture}
\end{equation}
where $x_0 \sim p_{\mathrm{data}}$ and $\epsilon \sim \mathcal{N}(0, I)$. Applying the Fourier transform yields
\begin{equation}
    \hat{x}_t = (1-t)\hat{x}_0 + t\hat{\epsilon}.
    \label{eq:fourier_mixture}
\end{equation}
Assuming independence between $\hat{x}_0$ and $\hat{\epsilon}$ with white noise spectrum, the expected PSD at spatial frequency $f$ is
\begin{equation}
    \mathbb{E}\!\left[\|\hat{x}_t\|_f^2\right] = (1-t)^2 S(f) + t^2 C_{\epsilon},
    \label{eq:psd_evolution}
\end{equation}
where $S(f)=\mathbb{E}\!\left[\|\hat{x}_0\|_f^2\right]$ is the clean data PSD and $C_{\epsilon}$ is a frequency-independent noise constant.

Each frequency band is governed by a competition between the clean signal and the noise floor. We define the signal-dominance ratio
\begin{equation}
    \rho(f,t)=\frac{(1-t)^2 S(f)}{t^2 C_{\epsilon}}.
    \label{eq:signal_ratio}
\end{equation}
Since $S(f)$ decreases with frequency, for any $f_l < f_h$ we have $\rho(f_l,t) > \rho(f_h,t)$, meaning lower-frequency bands become signal-dominant earlier. For visualization, we define the bounded recovery score
\begin{equation}
    P(f,t)=\frac{\rho(f,t)}{1+\rho(f,t)} \in [0,1],
    \label{eq:recovery_score}
\end{equation}
where larger values indicate greater signal dominance. Fig.~\ref{fig:spectral_dynamics} visualizes this quantity over normalized frequency and reverse denoising progress.

The critical transition time $t_c(f)$, at which signal and noise are equally strong ($\rho(f,t_c)=1$), is defined as:
\begin{equation}
    t_c(f)=\frac{\sqrt{S(f)}}{\sqrt{S(f)}+\sqrt{C_{\epsilon}}}.
    \label{eq:critical_time}
\end{equation}
Because $t_c(f)$ monotonically increases with $S(f)$, and $S(f)$ naturally decays at higher frequencies, the critical time for high-frequency components is strictly smaller (i.e., closer to $t=0$) than for low-frequency ones. In the reverse flow-matching process (from $t=1$ to $t=0$), this mathematical property dictates that low-frequency global layouts emerge early, while high-frequency micro-details are exclusively recoverable in the terminal stage. Combined with the empirical finding that RS imagery is heavily bottlenecked by high-frequency energy (Fig.~\ref{fig:empirical_psd}), this frequency-progressive behavior provides a compelling two-fold motivation for dynamic resolution promotion: RS images are both more reliant on high-frequency fidelity and more vulnerable to its suppression by static extrapolation. Driven by this crucial insight, we propose SHARP, a dynamic resolution promotion strategy designed to explicitly couple the extrapolation strength with this instantaneous spectral state.

\begin{algorithm}[t]
\caption{SHARP Inference}
\label{alg:sharp}
\begin{algorithmic}[1]
\Require Trained RS-FLUX model $\mathcal{F}$; text prompt $c$; target resolution $L_{\mathrm{target}}$; training resolution $L_{\mathrm{train}}$; scheduler coefficient $\alpha_s$; transition bounds $\alpha,\beta$; denoising timesteps $\{t_n\}_{n=0}^{N}$ with $t_0=1,\,t_N=0$
\Ensure Generated image $\hat{x}_0$ at resolution $L_{\mathrm{target}}$
\State $s \gets L_{\mathrm{target}} / L_{\mathrm{train}}$ \Comment{Resolution promotion factor}
\State $x_{t_0} \sim \mathcal{N}(0, I)$ \Comment{Initialize noise at target resolution}
\For{$n = 0$ \textbf{to} $N-1$}
    \State $\kappa_{rs}(t_n) \gets t_n / [\alpha_s - (\alpha_s - 1)\,t_n]$ \Comment{Rational Decay Scheduler}
    \State $\alpha_{t_n} \gets \alpha \cdot \kappa_{rs}(t_n)$;\quad $\beta_{t_n} \gets \beta \cdot \kappa_{rs}(t_n)$ \Comment{Dynamic bounds}
    \For{each RoPE dimension $i$}
        \State $r(i) \gets L_{\mathrm{target}} / \lambda_i$
        \State $\gamma \gets \mathrm{clamp}\!\bigl((r(i) - \alpha_{t_n}) / (\beta_{t_n} - \alpha_{t_n}),\, 0,\, 1\bigr)$ \Comment{Dynamic ramp}
        \State $h_{t_n}(\theta_i) \gets (1 - \gamma)\,\theta_i / s + \gamma\,\theta_i$ \Comment{Rescaled frequency}
    \EndFor
    \State Apply $\{h_{t_n}(\theta_i)\}_i$ to RoPE in $\mathcal{F}$
    \State $x_{t_{n+1}} \gets \mathrm{Denoise}(\mathcal{F},\, x_{t_n},\, t_n,\, c)$ \Comment{One denoising step}
\EndFor
\State $\hat{x}_0 \gets \mathrm{Decode}(x_{t_N})$ \Comment{VAE decode}
\State \Return $\hat{x}_0$
\end{algorithmic}
\end{algorithm}

\begin{figure*}[t]
    \centering
    \subfloat[Resolution distribution.]{\includegraphics[width=0.45\textwidth]{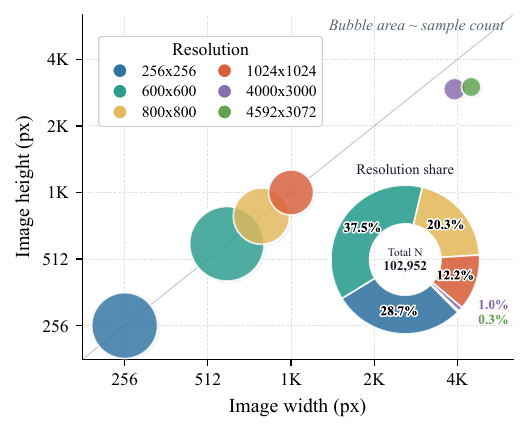}%
    \label{fig:dataset_a}}\hfill
    \subfloat[Corpus construction pipeline.]{\includegraphics[width=0.45\textwidth]{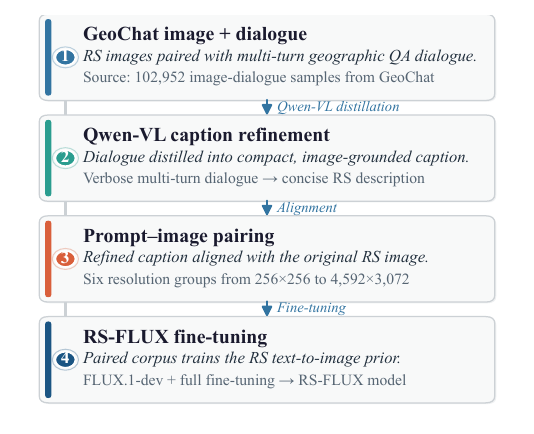}%
    \label{fig:dataset_b}}
    \caption{\textbf{Overview of the 102{,}952-sample RS training corpus.} (a)~Resolution distribution: bubble position and area denote dimensions and sample counts, respectively, with a composition summary in the inset. (b)~Corpus construction: GeoChat image-dialogue pairs are refined via Qwen-VL into descriptive captions for RS-FLUX fine-tuning.}
    \Description{Two-panel overview of the training corpus: a bubble chart showing the image resolution distribution, and a flowchart of the corpus construction pipeline converting GeoChat image-dialogue pairs into descriptive captions via Qwen-VL.}
    \label{fig:dataset_overview}
\end{figure*}
\subsection{SHARP: Dynamic Resolution Promotion}
\label{sec:sharp}

The spectral formulation in Sec.~\ref{sec:spectral_formulation} reveals a fundamental limitation of existing static extrapolation methods \cite{chen2023extending,bloc97ntk2023,peng2024yarn}. A time-invariant scaling rule is beneficial only during the early high-noise stage, where expanding the receptive field helps organize the global layout. However, it becomes highly detrimental in the late low-noise stage, where the exact same positional compression irreversibly suppresses the high-frequency components critical for RS detail recovery.

SHARP fundamentally resolves this bottleneck through a spectrum-aware, dynamically modulated promotion mechanism operating directly within the RoPE space. As illustrated in Fig.~\ref{fig:sharp_structure}, at each denoising step $t$, SHARP dynamically calibrates the RoPE frequency table via two tightly coupled modules: (i)~the \emph{Rational Decay Scheduler} (RDS), which governs the temporal evolution of the promotion strength in alignment with the frequency-progressive recovery order (Eq.~\ref{eq:critical_time}); and (ii)~the \emph{Dynamic Ramp Function}, which partitions the RoPE dimensions based on their effective spatial frequencies and applies differential extrapolation scaling across the frequency axis.

\textbf{Rational Decay Scheduler (RDS).}\enspace The RDS maps the normalized denoising timestep $t \in [0, 1]$ to a dynamic decay coefficient:
\begin{equation}
    \kappa_{rs}(t) = \frac{t}{\alpha_s - (\alpha_s - 1)t}, \quad \alpha_s \ge 1,
    \label{eq:rds}
\end{equation}
where $\alpha_s$ is the scheduler coefficient controlling the decay rate. This rational function is designed to smoothly decay from near-unity to zero. During early denoising steps ($t \approx 1$), $\kappa_{rs}(t) \approx 1$, maintaining strong positional promotion to establish macroscopic layouts. As generation progresses toward the terminal stage ($t \to 0$), $\kappa_{rs}(t)$ smoothly decays to $0$, explicitly releasing the model from excessive spatial compression when micro-textures are crystallizing.

\textbf{Dynamic Ramp Function.}\enspace We use the scheduled coefficient $\kappa_{rs}(t)$ to continuously modulate the transition boundaries that partition the frequency domain (as shown in Fig.~\ref{fig:sharp_structure}):
\begin{equation}
    \alpha_t = \alpha \cdot \kappa_{rs}(t), \quad \beta_t = \beta \cdot \kappa_{rs}(t),
    \label{eq:sharp_bounds}
\end{equation}
where $\alpha$ and $\beta$ are base threshold hyperparameters. These dynamic boundaries dictate the effective range of positional extrapolation at any instantaneous timestep $t$.

To determine how each RoPE dimension is processed, recall from Sec.~\ref{sec:rope_extrapolation} that the $i$-th feature dimension relies on a base frequency $\theta_i$. The corresponding spatial wavelength of this rotation is mathematically defined as $\lambda_i = 2\pi / \theta_i$. Let $s=L_{\mathrm{target}}/L_{\mathrm{train}}$ denote the target resolution promotion factor. We define the normalized spatial frequency ratio for the $i$-th dimension as:
\begin{equation}
    r(i)=\frac{L_{\mathrm{target}}}{\lambda_i} = \frac{L_{\mathrm{target}} \theta_i}{2\pi}.
    \label{eq:r_def}
\end{equation}
This ratio $r(i)$ elegantly quantifies how many full wavelengths fit within the target sequence length. SHARP then rescales each RoPE frequency through a time-dependent interpolation:
\begin{equation}
    h_t(\theta_i)=\bigl(1-\gamma(r(i),t)\bigr)\frac{\theta_i}{s} + \gamma(r(i),t)\,\theta_i,
    \label{eq:sharp_ht}
\end{equation}
where $\gamma(r,t)$ is the Dynamic Ramp Function defined by the boundaries $\alpha_t$ and $\beta_t$:
\begin{equation}
    \gamma(r,t)=
    \begin{cases}
        0, & r < \alpha_t, \\[4pt]
        \dfrac{r-\alpha_t}{\beta_t-\alpha_t}, & \alpha_t \le r \le \beta_t, \\[4pt]
        1, & r > \beta_t.
    \end{cases}
    \label{eq:sharp_gamma}
\end{equation}

This mathematical design directly operationalizes the generative physics derived in Sec.~\ref{sec:spectral_formulation}. Equation~\ref{eq:sharp_ht} demonstrates that low-frequency modes ($r < \alpha_t$) receive full linear extrapolation ($\theta_i/s$), high-frequency modes ($r > \beta_t$) are left perfectly uncompressed ($\theta_i$), and intermediate modes undergo smooth interpolation. 

Crucially, because $\alpha_t$ and $\beta_t$ decay with $\kappa_{rs}(t)$, the behavior is \emph{highly dynamic}. In the early high-noise stage, the transition band is exceptionally wide, enforcing uniform spatial promotion across most frequencies to guarantee layout coherence. As denoising approaches the terminal detail-refinement phase, the boundaries $\alpha_t$ and $\beta_t$ shrink toward zero. This aggressively shifts the vast majority of frequency modes into the uncompressed regime ($\gamma=1$), thereby meticulously preserving the crisp high-frequency content essential for photorealistic RS imagery.

Algorithm~\ref{alg:sharp} summarizes the complete SHARP inference procedure.

\begin{figure*}[t]
    \centering
    \includegraphics[width=0.95\textwidth]{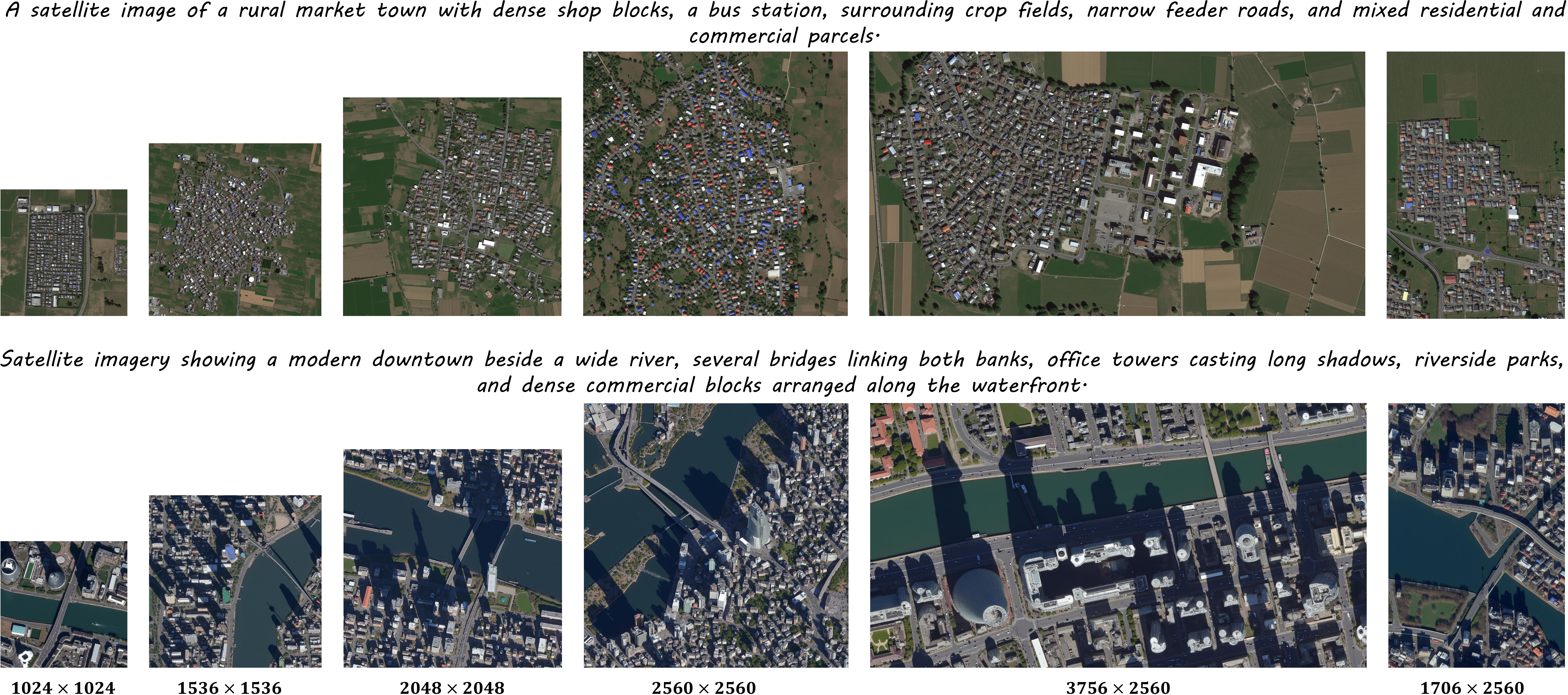}
    \caption{\textbf{Multi-scale generation from a single prompt.} Each row demonstrates SHARP's generation across six diverse resolutions and aspect ratios (from native $1024{\times}1024$ up to $3756{\times}2560$). SHARP maintains strict global semantic consistency across all scales while progressively crystallizing finer RS details at higher resolutions. This visual coherence validates that our dynamic RoPE scheduling successfully decouples macro-layout formation from high-frequency detail recovery.}
    \label{fig:qualitative_many_ratio}
    \Description{A grid showing remote sensing images generated by SHARP across six different resolutions, demonstrating consistent spatial layouts and progressively sharper details.}
\end{figure*}

\textbf{Multi-scale Compatibility.}\enspace 
SHARP is independent of absolute pixel dimensions since it operates on normalized timesteps $t \in [0,1]$ and dimensionless ratios $r(i)$: all hyperparameters ($\alpha_s, \alpha, \beta$) remain fixed across targets, and only the promotion factor $s = L_{\mathrm{target}}/L_{\mathrm{train}}$ is updated, computed per spatial axis for anisotropic targets with the same ramp partitioning the corresponding RoPE dimensions. This "one-set-fits-all" configuration enables robust inference across diverse resolutions without per-resolution tuning, as validated in Sec.~\ref{sec:experiments}.

\section{Experiments}
\label{sec:experiments}

\subsection{Experimental Setup}

\subsubsection{Training Data Construction}

We construct a domain-specialized corpus from the GeoChat dataset \cite{kuckreja2024geochat}, which provides RS images with multi-turn dialogues. Since the question-answer format is poorly suited to caption-style supervision, we use Qwen-VL \cite{bai2023qwen} to convert each sample into a compact prompt-image pair (Fig.~\ref{fig:dataset_b}), yielding 102{,}952 pairs for RS-FLUX fine-tuning. The corpus spans multiple resolutions (Fig.~\ref{fig:dataset_a}), promoting robustness to varying spatial extents.

\begin{figure}[t]
    \centering
    \includegraphics[width=1.0\linewidth]{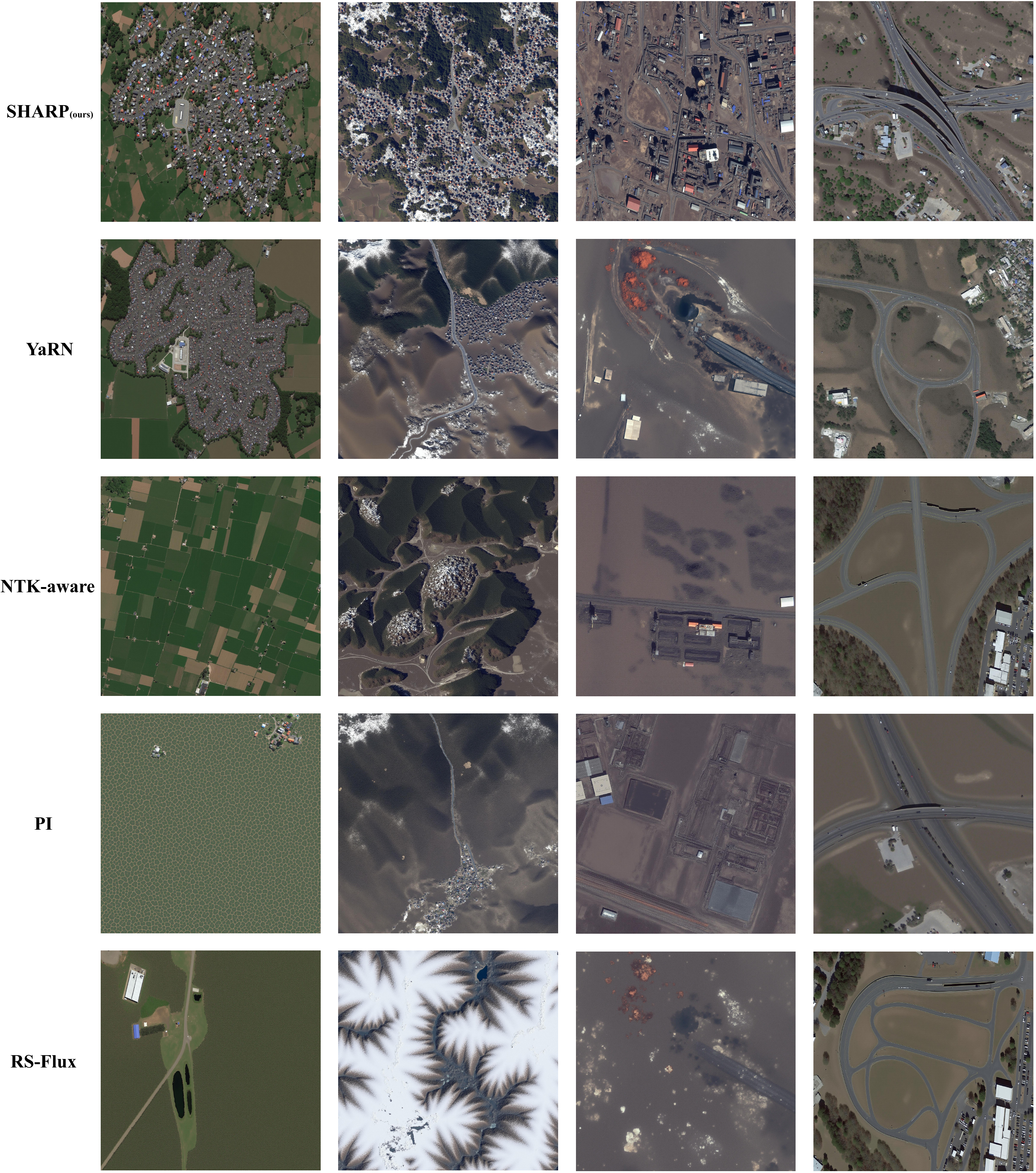}
    \caption{\textbf{Qualitative comparison at $2048{\times}2048$ resolution.} \textbf{Please zoom in for better detail visualization.} Unlike static baselines that suffer from catastrophic over-smoothing of fine structures, SHARP consistently preserves crisp high-frequency features (e.g., dense buildings and road topologies) while maintaining superior global structural coherence.}
    \Description{A grid of remote sensing images comparing SHARP with static baselines at 2048 by 2048 resolution, where SHARP preserves crisp buildings and road structures while baselines appear over-smoothed.}
    \label{fig:qualitative_2048}
\end{figure}

\subsubsection{Evaluation Protocol}
We construct a benchmark of 100 GPT-5.4 generated prompts \cite{singh2025openai} covering diverse RS scenes. We evaluate generation quality using three reference-free metrics: CLIP Score \cite{hessel2021clipscore} (semantic consistency), Aesthetic Score \cite{schuhmann2022laion} (perceptual quality), and HPSv2 \cite{wu2023human} (human preference).

\begin{table}[t]
    \centering
    \caption{\textbf{Native-resolution comparison at $1024\times1024$.} Higher is better.}
    \label{tab:rsflux_native}
    \begin{tabular}{lccc}
        \toprule
        Model & CLIP$\uparrow$ & Aes$\uparrow$ & HPSv2$\uparrow$ \\
        \midrule
        FLUX (vanilla) & 25.53 & 5.18 & 0.236 \\
        \rowcolor{ourmethodbg}
        RS-FLUX (ours) & \textbf{28.62} & \textbf{5.97} & \textbf{0.283} \\
        \bottomrule
    \end{tabular}
\end{table}

\subsubsection{Multi-Scale Promotion Setting}
To evaluate generation beyond the $1024\times1024$ training resolution, we test on three isotropic targets ($1536^2$, $2048^2$, $2560^2$) and three anisotropic targets ($2048\times1536$, $2560\times1920$, $3072\times2048$). All methods generate images from the same 100 prompts under identical random seeds.

\subsubsection{Implementation Details}
RS-FLUX is fine-tuned from the $1024^2$ FLUX checkpoint using AdamW ($\beta_1{=}0.9, \beta_2{=}0.999$, weight decay $0.01$) with a $10^{-5}$ learning rate and cosine annealing. Training spans 10K steps with bf16 precision and an effective batch size of 64 (8$\times$NVIDIA A6000 GPUs). For fair comparison, all inference methods share the same backbone, Euler sampler (28 steps, guidance scale 3.5), and seeds; only the RoPE adaptation strategy varies. SHARP hyperparameters ($\alpha_s{=}3, \alpha{=}1, \beta{=}32$) follow the ablation results in Sec.~\ref{sec:ablation}.
\subsection{RS-FLUX Validation at Native Resolution}
\label{sec:rsflux_validation}

\begin{table*}[t]
    \centering
    \caption{\textbf{Multi-scale quantitative comparison under isotropic and anisotropic promotion.} Higher is better for all metrics. The upper block evaluates isotropic promotion; the lower block evaluates anisotropic promotion. ``Overall Avg.'' averages all six resolutions.}
    \label{tab:multiscale_main}
    \resizebox{\textwidth}{!}{
    \begin{tabular}{lcccccccccccc}
        \toprule
        \rowcolor{tableblockbg}
        \multicolumn{13}{c}{\textbf{Isotropic Promotion}} \\
        \midrule
        \multirow{2}{*}{Method} &
        \multicolumn{3}{c}{$1536\times1536$} &
        \multicolumn{3}{c}{$2048\times2048$} &
        \multicolumn{3}{c}{$2560\times2560$} &
        \multicolumn{3}{c}{Isotropic Avg.} \\
        \cmidrule(lr){2-4} \cmidrule(lr){5-7} \cmidrule(lr){8-10} \cmidrule(lr){11-13}
        & CLIP$\uparrow$ & Aes$\uparrow$ & HPSv2$\uparrow$
        & CLIP$\uparrow$ & Aes$\uparrow$ & HPSv2$\uparrow$
        & CLIP$\uparrow$ & Aes$\uparrow$ & HPSv2$\uparrow$
        & CLIP$\uparrow$ & Aes$\uparrow$ & HPSv2$\uparrow$ \\
        \midrule
        RS-FLUX (baseline) & 26.84 & 5.48 & 0.249 & 26.21 & 5.36 & 0.242 & 25.57 & 5.25 & 0.236 & 26.21 & 5.36 & 0.242 \\
        PI & 27.18 & 5.56 & 0.255 & 26.73 & 5.46 & 0.248 & 26.09 & 5.33 & 0.241 & 26.67 & 5.45 & 0.248 \\
        NTK-aware & 27.44 & 5.63 & 0.261 & 27.02 & 5.54 & 0.255 & 26.46 & 5.42 & 0.248 & 26.97 & 5.53 & 0.255 \\
        YaRN & 27.71 & 5.78 & 0.269 & 27.31 & 5.69 & 0.263 & 26.78 & 5.57 & 0.257 & 27.27 & 5.68 & 0.263 \\
        \rowcolor{ourmethodbg}
        SHARP (ours) & \textbf{28.03} & \textbf{5.89} & \textbf{0.276} & \textbf{27.72} & \textbf{5.82} & \textbf{0.271} & \textbf{27.28} & \textbf{5.71} & \textbf{0.266} & \textbf{27.68} & \textbf{5.81} & \textbf{0.271} \\
        \midrule
        \rowcolor{tableblockbg}
        \multicolumn{13}{c}{\textbf{Anisotropic Promotion}} \\
        \midrule
        \multirow{2}{*}{Method} &
        \multicolumn{3}{c}{$2048\times1536$} &
        \multicolumn{3}{c}{$2560\times1920$} &
        \multicolumn{3}{c}{$3072\times2048$} &
        \multicolumn{3}{c}{Overall Avg.} \\
        \cmidrule(lr){2-4} \cmidrule(lr){5-7} \cmidrule(lr){8-10} \cmidrule(lr){11-13}
        & CLIP$\uparrow$ & Aes$\uparrow$ & HPSv2$\uparrow$
        & CLIP$\uparrow$ & Aes$\uparrow$ & HPSv2$\uparrow$
        & CLIP$\uparrow$ & Aes$\uparrow$ & HPSv2$\uparrow$
        & CLIP$\uparrow$ & Aes$\uparrow$ & HPSv2$\uparrow$ \\
        \midrule
        RS-FLUX (baseline) & 26.48 & 5.42 & 0.246 & 25.94 & 5.31 & 0.239 & 25.38 & 5.20 & 0.233 & 26.07 & 5.34 & 0.241 \\
        PI & 26.96 & 5.50 & 0.252 & 26.42 & 5.40 & 0.246 & 25.88 & 5.28 & 0.239 & 26.54 & 5.42 & 0.247 \\
        NTK-aware & 27.25 & 5.58 & 0.259 & 26.74 & 5.48 & 0.252 & 26.19 & 5.36 & 0.245 & 26.85 & 5.50 & 0.253 \\
        YaRN & 27.53 & 5.72 & 0.267 & 27.04 & 5.62 & 0.261 & 26.50 & 5.50 & 0.254 & 27.15 & 5.65 & 0.262 \\
        \rowcolor{ourmethodbg}
        SHARP (ours) & \textbf{27.86} & \textbf{5.84} & \textbf{0.274} & \textbf{27.40} & \textbf{5.75} & \textbf{0.269} & \textbf{26.93} & \textbf{5.64} & \textbf{0.263} & \textbf{27.54} & \textbf{5.78} & \textbf{0.270} \\
        \bottomrule
    \end{tabular}}
\end{table*}

Before evaluating resolution promotion strategies, we verify that domain-specialized fine-tuning improves generation quality at the native $1024\times1024$ resolution. Table~\ref{tab:rsflux_native} compares vanilla FLUX with RS-FLUX on the same 100-prompt benchmark.

RS-FLUX improves over vanilla FLUX by +3.09 CLIP, +0.79 Aes, and +0.047 HPSv2. The largest gain appears on CLIP Score (+12.1\% relative), indicating that full fine-tuning substantially narrows the semantic gap between generated outputs and RS-specific prompts---vanilla FLUX frequently misinterprets aerial terminology. The Aesthetic Score and HPSv2 improvements further confirm that RS-FLUX produces visually coherent aerial scenes with realistic color palettes and spatial layouts, establishing a stronger starting point for resolution promotion.

\subsection{Quantitative Comparison}

\begin{table*}[t]
    \caption{\textbf{Ablation studies at $2048\times2048$.} Higher is better. CL\,=\,CLIP Score, Aes\,=\,Aesthetic Score, HP\,=\,HPSv2. (a)~Component contributions (FT\,=\,fine-tuning, SH\,=\,SHARP). (b)~Scheduler form. (c)~Scheduler coefficient $\alpha_s$. (d)~Transition bounds $(\alpha,\beta)$.}
    \label{tab:ablation}
    \centering
    \small
    \begin{minipage}[t]{0.235\textwidth}
        \centering
        {\footnotesize\textbf{(a)} Component}\\[2pt]
        \begin{tabular}{ccccc}
            \toprule
            FT & SH & CL & Aes & HP \\
            \midrule
                  &       & 24.83 & 5.08 & .228 \\
                  & \checkmark & 26.15 & 5.48 & .253 \\
            \checkmark &       & 26.21 & 5.36 & .242 \\
            \rowcolor{ourmethodbg}
            \checkmark & \checkmark & \textbf{27.72} & \textbf{5.82} & \textbf{.271} \\
            \bottomrule
        \end{tabular}
    \end{minipage}
    \hfill
    \begin{minipage}[t]{0.255\textwidth}
        \centering
        {\footnotesize\textbf{(b)} Scheduler form}\\[2pt]
        \begin{tabular}{lccc}
            \toprule
            Schedule & CL & Aes & HP \\
            \midrule
            Static & 27.35 & 5.70 & .264 \\
            Linear & 27.48 & 5.74 & .266 \\
            Cosine & 27.58 & 5.77 & .268 \\
            \rowcolor{ourmethodbg}
            Rational & \textbf{27.72} & \textbf{5.82} & \textbf{.271} \\
            \bottomrule
        \end{tabular}
    \end{minipage}
    \hfill
    \begin{minipage}[t]{0.21\textwidth}
        \centering
        {\footnotesize\textbf{(c)} Coefficient $\alpha_s$}\\[2pt]
        \begin{tabular}{cccc}
            \toprule
            $\alpha_s$ & CL & Aes & HP \\
            \midrule
            1 & 27.48 & 5.74 & .266 \\
            2 & 27.61 & 5.78 & .269 \\
            \rowcolor{ourmethodbg}
            3 & \textbf{27.72} & \textbf{5.82} & \textbf{.271} \\
            4 & 27.55 & 5.75 & .267 \\
            \bottomrule
        \end{tabular}
    \end{minipage}
    \hfill
    \begin{minipage}[t]{0.25\textwidth}
        \centering
        {\footnotesize\textbf{(d)} Bounds $(\alpha,\beta)$}\\[2pt]
        \begin{tabular}{ccccc}
            \toprule
            $\alpha$ & $\beta$ & CL & Aes & HP \\
            \midrule
            1 & 16 & 27.45 & 5.71 & .264 \\
            \rowcolor{ourmethodbg}
            1 & 32 & \textbf{27.72} & \textbf{5.82} & \textbf{.271} \\
            2 & 32 & 27.61 & 5.77 & .268 \\
            2 & 64 & 27.38 & 5.66 & .261 \\
            \bottomrule
        \end{tabular}
    \end{minipage}
\end{table*}

Table~\ref{tab:multiscale_main} compares SHARP with four baselines across six resolutions. The results show a consistent ranking: RS-FLUX (no RoPE adaptation) $<$ PI $<$ NTK-aware $<$ YaRN $<$ SHARP. Although PI, NTK-aware, and YaRN progressively improve the frequency treatment, they all remain static during denoising. By introducing dynamic scheduling, SHARP consistently outperforms the strongest static baseline.

\textbf{Overall performance.} SHARP achieves 27.54 CLIP, 5.78 Aes, and 0.270 HPSv2 on average, surpassing the strongest baseline YaRN by +0.39 / +0.13 / +0.008 and RS-FLUX by +1.47 / +0.44 / +0.029.

\textbf{Isotropic promotion.} SHARP improves over YaRN from 27.27 / 5.68 / 0.263 to 27.68 / 5.81 / 0.271. The margin \emph{widens} with the promotion factor: +0.32 / +0.11 / +0.007 at $1536^2$ versus +0.50 / +0.14 / +0.009 at $2560^2$. This widening corroborates the spectrum-aware design: as the promotion factor grows, static methods suffer increasingly severe high-frequency suppression, whereas SHARP's dynamic scheduling compensates accordingly. From $1536^2$ to $2560^2$, SHARP's metrics degrade by 0.75 / 0.18 / 0.010 versus 0.93 / 0.21 / 0.012 for YaRN, indicating slower quality decay under aggressive promotion.

\textbf{Anisotropic promotion.} SHARP leads at all anisotropic settings, with the largest margins at the most challenging $3072\times2048$ (26.93 / 5.64 / 0.263 vs.\ 26.50 / 5.50 / 0.254), handling differing per-axis promotion factors gracefully through axis-independent RoPE rescaling via the dimensionless ratio $r(i)$.

\subsection{Qualitative Comparison}

Fig.~\ref{fig:qualitative_2048} presents qualitative results at $2048\times2048$. Static baselines suffer from severe structural collapse, frequently degrading into hallucinated or overly smoothed homogeneous regions (e.g., the dense residential and industrial scenes); YaRN roughly preserves macroscopic layouts but heavily blurs detail-critical regions. In contrast, SHARP retains both global coherence and high-frequency details, synthesizing crisp building footprints, complex terrain textures, and multi-level highway topologies.

Fig.~\ref{fig:qualitative_many_ratio} demonstrates multi-scale consistency. Using the same prompt, SHARP generates RS images across six resolutions from $1024\times1024$ to $3756\times2560$. Coherent spatial composition is maintained at all scales while finer details---vehicle contours, building boundaries, and road textures---progressively emerge as resolution increases. Spatial relationships remain stable even as the canvas expands and the aspect ratio changes, confirming that the Rational Decay Scheduler preserves global layout while the Dynamic Ramp Function progressively releases high-frequency modes.

\subsection{Ablation Studies}
\label{sec:ablation}

\begin{table}[t]
    \centering
    \caption{\textbf{Inference cost (seconds per image).} Peak GPU memory is identical across all methods.}
    \label{tab:efficiency}
    \begin{tabular}{lccc}
        \toprule
        Method & $1536^2$ & $2048^2$ & $2560^2$ \\
        \midrule
        RS-FLUX   & 66.7 & 130.8 & 235.4 \\
        PI        & 67.1 & 131.2 & 237.1 \\
        NTK-aware & 68.8 & 133.1 & 238.9 \\
        YaRN      & 68.4 & 132.3 & 238.2 \\
        \rowcolor{ourmethodbg}
        SHARP     & 69.1 & 134.1 & 243.7 \\
        \bottomrule
    \end{tabular}
\end{table}

All ablations run at $2048\times2048$ on the 100-prompt benchmark. Table~\ref{tab:ablation}(a) confirms that fine-tuning and SHARP are complementary: each independently improves over vanilla FLUX, and their combination yields the best result (+2.89 CLIP). SHARP alone on vanilla FLUX (26.15 CLIP) already rivals RS-FLUX without SHARP (26.21 CLIP)---fine-tuning enriches the semantic prior while SHARP addresses the orthogonal positional OOD problem.

Panel~(b) compares four scheduling strategies, all satisfying $\kappa(1){=}1$ and $\kappa(0){=}0$ with identical bounds $(\alpha,\beta){=}(1,32)$. Even the simplest linear schedule outperforms the static baseline by +0.13 CLIP. The rational form performs best: it releases most of the positional compression soon after the global layout stabilizes and then decays smoothly to zero, best matching the frequency-progressive recovery timeline.

Panel~(c) reveals a unimodal optimum at $\alpha_s{=}3$: at $\alpha_s{=}1$ (linear decay), promotion decreases too gradually, leaving residual late-stage compression, while $\alpha_s{=}4$ releases promotion too early, weakening mid-stage layout formation. The $\alpha_s{=}1$ result matches the linear entry in panel~(b).

Panel~(d) confirms $(\alpha,\beta){=}(1,32)$ as optimal, coinciding with YaRN's default ramp parameters \cite{peng2024yarn}, indicating that the frequency-domain partition is robust across modalities and SHARP's gains stem from the orthogonal temporal dimension. Too small $\beta$ ($16$) lets most modes bypass promotion, while too large $\beta$ ($64$) interpolates nearly all modes uniformly.

\subsection{Computational Efficiency}
\label{sec:efficiency}

SHARP modifies only the RoPE frequency computation at each step. Table~\ref{tab:efficiency} reports inference time on a single A6000 GPU: SHARP adds only 2.5\%--3.6\% overhead relative to vanilla RS-FLUX across all tested resolutions. The extra cost---element-wise arithmetic over the RoPE dimensions per step (Eq.~\ref{eq:rds}--\ref{eq:sharp_gamma})---is negligible compared with the self-attention and feed-forward computations dominating each pass, and peak GPU memory is unchanged.

\section{Conclusion}
\label{sec:conclusion}

We present SHARP, a training-free resolution promotion framework for large-scale RS synthesis. Replacing static extrapolation with a Rational Decay Scheduler $\kappa_{rs}(t)$, SHARP preserves global layouts and dense high-frequency details, consistently outperforming static baselines with negligible ($<4\%$) overhead. Future work includes content-adaptive schedules and sparse attention for extreme resolutions.

\begin{acks}
This work was supported in part by the National Natural Science Foundation of China under Grant 62471394.
\end{acks}

\bibliographystyle{ACM-Reference-Format}
\bibliography{egbib}

\end{document}

%% file: settings.tex
\definecolor{pinkkk}{RGB}{237,218,244}
\definecolor{pinkkk1}{RGB}{248,235,249}
\definecolor{redd}{RGB}{205,143,143}
\definecolor{grayyy}{RGB}{242,242,242}
\definecolor{greenn}{RGB}{124,191,51}
\definecolor{lightergray}{gray}{0.92} 
\definecolor{tableblockbg}{RGB}{234,239,245}
\definecolor{ourmethodbg}{RGB}{232,243,233}